\documentclass[letterpaper]{article} 
\usepackage{aaai25}  
\usepackage{times}  
\usepackage{helvet}  
\usepackage{courier}  
\usepackage[hyphens]{url}  
\usepackage{graphicx} 
\usepackage{natbib}  
\usepackage{caption} 
\usepackage{algorithm}
\usepackage{multirow}%
\usepackage{amsmath,amssymb,amsfonts}%
\usepackage{amsthm}%
\usepackage{mathrsfs}%
\usepackage{xcolor}%
\usepackage{textcomp}%
\usepackage{manyfoot}%
\usepackage{booktabs}%
\usepackage{algorithm}%
\usepackage{algorithmicx}%
\usepackage{algpseudocode}%
\usepackage{listings}%
\usepackage{pifont}
\usepackage{svg} 

\usepackage{color, colortbl}
\usepackage{multirow}
\usepackage{makecell}
\usepackage{comment} 

\definecolor{LightCyan}{rgb}{0.88,0.95,1}
\definecolor{mygreen}{HTML}{007F7F}
\definecolor{myred}{HTML}{CC0000}
\definecolor{myblue}{RGB}{0, 71, 171}
\definecolor{gold}{rgb}{0.75, 0.5, 0.25}

\usepackage{newfloat}
\usepackage{listings}
\DeclareCaptionStyle{ruled}{labelfont=normalfont,labelsep=colon,strut=off} 
\floatstyle{ruled}
\newfloat{listing}{tb}{lst}{}
\floatname{listing}{Listing}
\nocopyright 

\title{\textcolor{mygreen}{Any}\textcolor{myred}{Design}: Versatile Area Fashion Editing via Mask-Free Diffusion}
\author {
    Yunfang Niu\textsuperscript{\rm 1,\rm 2},
    Dong Yi\textsuperscript{\rm 1,\rm 2,\rm 3},
    Lingxiang Wu\textsuperscript{\rm 1,\rm 2,\rm 3},
    Jie Peng\textsuperscript{\rm 1,\rm 4},
    Jinqiao Wang\textsuperscript{\rm 1,\rm 2,\rm 3}
}
\affiliations {
    \textsuperscript{\rm 1}Foundation Model Research Center, Institute of Automation, Chinese Academy of Sciences \\
    \textsuperscript{\rm 2} School of Artificial Intelligence, University of Chinese Academy of Sciences \\
    \textsuperscript{\rm 3}Wuhan AI Research \\
    \textsuperscript{\rm 4}Wuhan University \\
}
\usepackage{bibentry}

\begin{document}

\twocolumn[{
\renewcommand\twocolumn[1][]{#1}
\maketitle
\begin{center}
    \captionsetup{type=figure}
    \includegraphics[width=\textwidth]{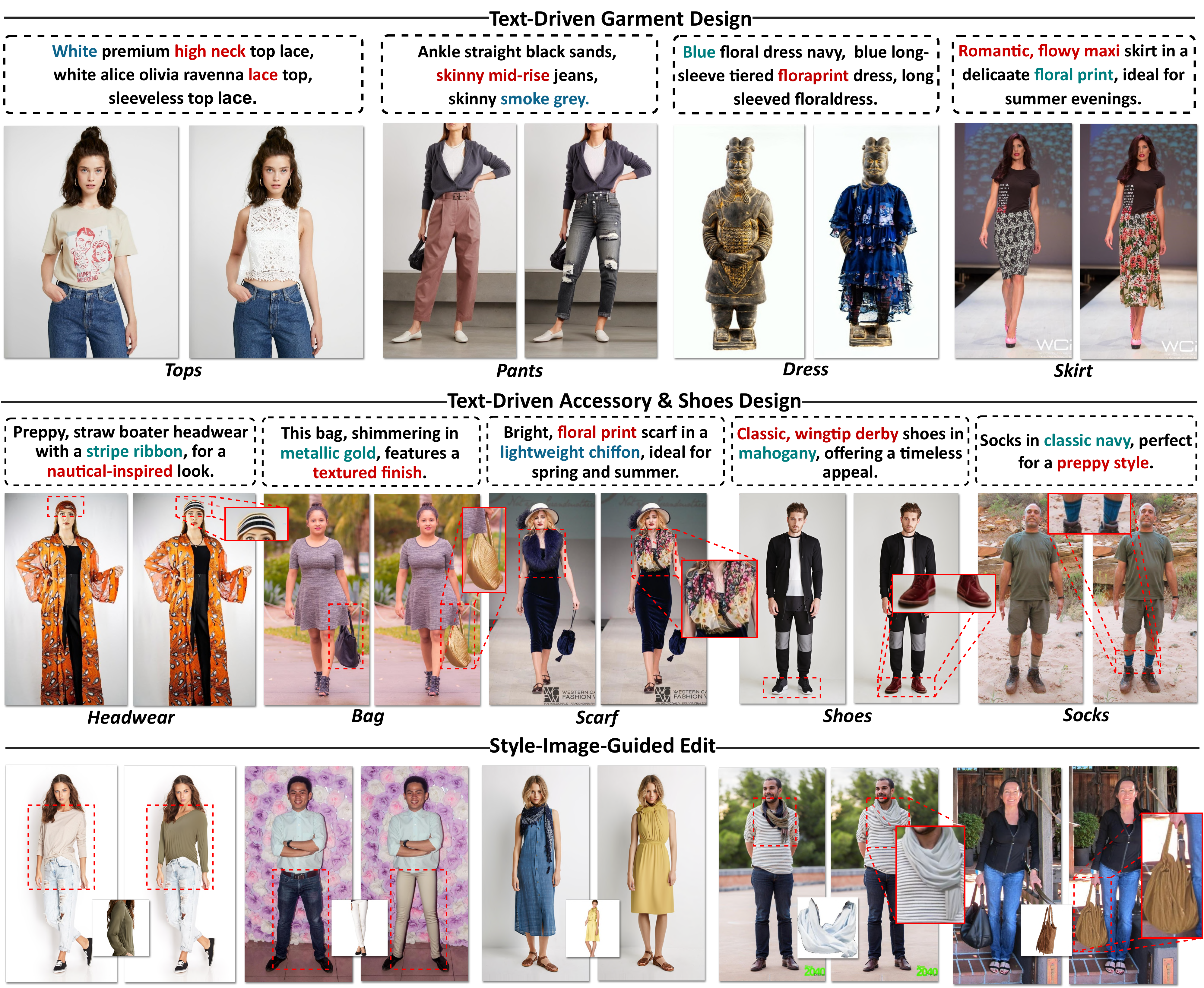}
    \captionof{figure}{Fashion Editing with AnyDesign. Our model adapts to various settings and edits a wide range of apparel categories using flexible prompts.}
    \label{fig:head_image}
\end{center}
}]

\clearpage

\begin{abstract}
Fashion image editing aims to modify a person's appearance based on a given instruction. Existing methods require auxiliary tools like segmenters and keypoint extractors, lacking a flexible and unified framework. Moreover, these methods are limited in the variety of clothing types they can handle, as most datasets focus on people in clean backgrounds and only include generic garments such as tops, pants, and dresses. These limitations restrict their applicability in real-world scenarios. In this paper, we first extend an existing dataset for human generation to include a wider range of apparel and more complex backgrounds. This extended dataset features people wearing diverse items such as tops, pants, dresses, skirts, headwear, scarves, shoes, socks, and bags. Additionally, we propose \textbf{AnyDesign}, a diffusion-based method that enables mask-free editing on versatile areas. Users can simply input a human image along with a corresponding prompt in either text or image format. Our approach incorporates Fashion DiT, equipped with a Fashion-Guidance Attention (FGA) module designed to fuse explicit apparel types and CLIP-encoded apparel features. Both Qualitative and quantitative experiments demonstrate that our method delivers high-quality fashion editing and outperforms contemporary text-guided fashion editing methods.
\end{abstract}

%

\section{Introduction}

Fashion image editing, a challenging yet innovative task, enables users to change clothing through prompt guidance, having wide applications in domains such as fashion design~\cite{sorger2017fundamentals} and E-commerce~\cite{jain2021overview}. This task can be categorized into image-based~\cite{han2018viton,choi2021viton,morelli2022dress} and text-based~\cite{pernuvs2023fice,baldrati2023multimodal} editing. The former covers the virtual try-on methods, which involve transferring the garment image onto the target person. The latter involves fashion image editing with text descriptions, which allows fashion designers to manipulate the model image by various natural language instructions. 

Successful E-commerce applications should allow users to upload their full-body shots in any background and modify the image with any apparel in image or text formats. However, existing methods typically require auxiliary information like dedicated masks~\cite{choi2021viton} and human poses~\cite{cao2017realtime,guler2018densepose}. These auxiliary tools face significant challenges in real-world applications, especially when the user's full-body shot is in an arbitrary pose and background. Additionally, existing methods can handle limited clothing types, such as tops, pants, and dresses. This is because public fashion editing datasets like VITON-HD~\cite{han2018viton} and Dresscode~\cite{morelli2022dress} only focus on mainstream clothing items and people in clean backgrounds. 

To address these challenges, we introduce \textbf{AnyDesign}, a mask-free framework for human fashion editing that allows users to edit versatile areas, including tops, pants, dresses, shoes, and accessories. 
Although fashion designers often edit by adding text inspiration to a human image, there are far more paired images of people with apparel than detailed text descriptions available. To leverage this, we utilize both human images paired with apparel images and those with text descriptions through the use of OpenFashionClip~\cite{cartella2023openfashionclip}. This enables AnyDesign to perform versatile editing, including text-driven garment design, text-driven accessory design, and style-image guided editing, as illustrated in Fig.~\ref{fig:head_image}.

We start our work by extending a dataset. SHHQ~\cite{fu2022stylegan} is a dataset designed for human generation tasks, featuring a wide range of human bodies with well-labeled attributes and diverse garment categories. However, it is not suitable for direct use in the human fashion editing task due to the absence of key components such as apparel-agnostic images, human densepose, and guidance prompts. In this paper, we introduce a dataset extension method, extending the SHHQ dataset for versatile human fashion editing in the wild, referred as SSHQe. Compared with the public datasets VITON-HD~\cite{choi2021viton} and Dresscode~\cite{morelli2022dress}, our extended dataset contains people in complex backgrounds with nine kinds of clothes and accessories \textit{(i.e., tops, pants, dresses, skirts, headwear, scarves, shoes, socks, bags)}. We also provide agnostic images and guidance prompts, applying different removal strategies for various apparel categories, along with dedicated human keypoints and denseposes.

On the extended dataset, an innovative mask-free framework is introduced for human fashion editing. The framework contains two stages. The first stage involves a mask-based diffusion model to generate pseudo samples. The second stage transitions to a mask-free diffusion model, which is employed during the inference phase. For the apparel encoder, we fully leverage the cross-modal capabilities of the CLIP-like features. Utilizing the OpenFashionClip~\cite{cartella2023openfashionclip} pre-trained on large-scale fashion datasets~\cite{han2017automatic,rostamzadeh2018fashion,guo2019imaterialist,wu2021fashion}, our method supports both text-guided and style-image-guided clothing editing as both the 
textual and visual features are semantically aligned. For the diffusion model backbone, we design a Fashion Diffusion Transformer (Fashion DiT), where a novel fashion-guidance attention (FGA) is proposed to integrate explicit apparel types and the encoded apparel features. By specifying the apparel types and desired modifications, the framework automatically identifies the target areas and completes the synthesis end-to-end. To validate the effectiveness of the proposed method, we conduct comprehensive experiments on both our newly extended dataset and the most widely used datasets for human fashion editing.

\begin{figure*}[t]
    \centering
    \includegraphics[width=\linewidth]{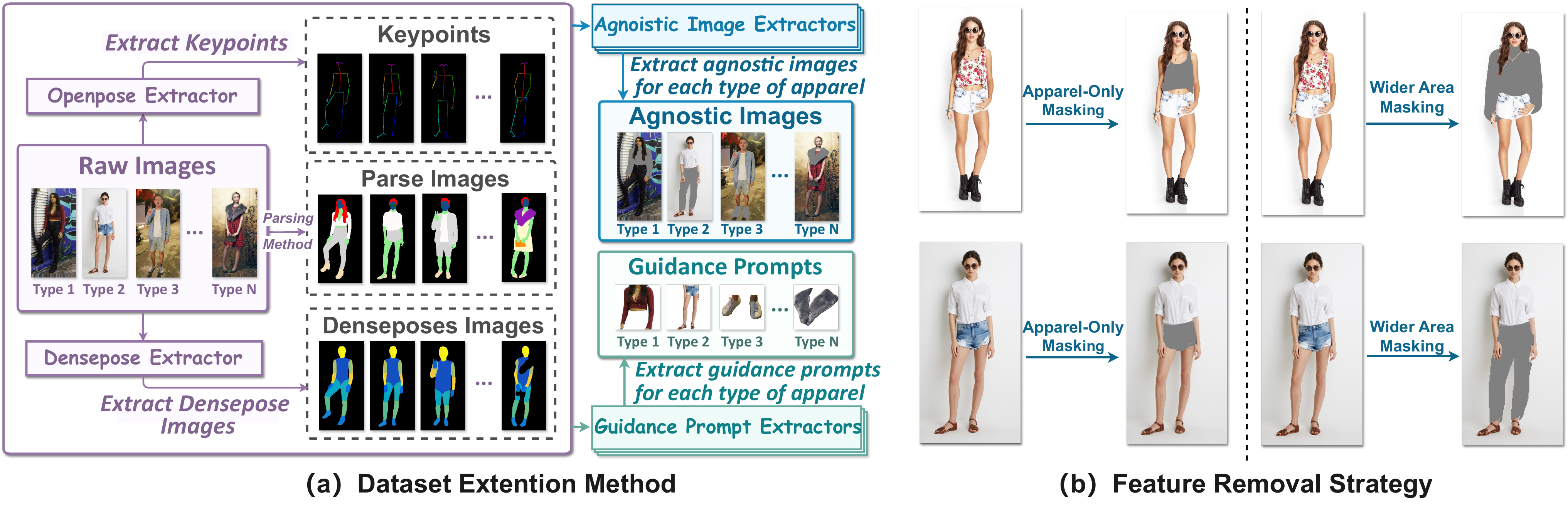}
    \caption{(a) The Dataset Extension Method. We extract keypoints and densepose information using existing methods. Then, apparel-specific extractors are designed to create agnostic images and guidance prompts. (b) Different feature removal strategies.}
    \label{fig:agnostic_generation}
\end{figure*} 

The contributions of this paper can be summarized as follows:
\begin{itemize}
    \item We introduce a dataset extension method and first reproduce an extended fashion editing dataset featuring people in complex backgrounds and nine categories of apparel.  
    \item We introduce an innovative mask-free diffusion framework for human fashion editing. In the framework, we propose a Fashion DiT, incorporating a fashion guidance attention module that fuses CLIP-like features with apparel types.
    \item Experiments validate our model's efficiency on various datasets, demonstrating notable performance improvements and extensive application potential.
\end{itemize}

\section{Related Work}
\label{sec:P2}
\subsubsection{Text-to-Image Synthesis.} This challenging task involves synthesizing a target image based on a given textual prompt. Initially, this task primarily relied on Generative Adversarial Networks (GANs)~\cite{zhang2017stackgan, zhang2021cross, tao2022df}, while more recent work has predominantly adopted diffusion-based models~\cite{nichol2021glide, rombach2022high, saharia2022photorealistic}.
In the fashion image synthesis, Fashion-Gen~\cite{rostamzadeh2018fashion} firstly constructed a large-scale dataset and utilized StackGAN~\cite{zhang2017stackgan} for text-to-fashion image generation. FashionG~\cite{jiang2021deep} incorporated global and local style losses to generate images with multiple styles. UFS-Net~\cite{wu2023ufs} enabled sketch-guided garment generation under an unsupervised learning paradigm. In diffusion-based approaches, a fashion attribute editing framework~\cite{kong2023leveraging} was proposed, exploring the classifier-guided diffusion~\cite{dhariwal2021diffusion}. Then, SGDiff~\cite{sun2023sgdiff} innovatively used a diffusion model to synthesize garment images with style guidance.

\subsubsection{Text-Driven Fashion Editing.} This line of work focuses on the precise editing of given images via textual description. Extensive research has been conducted on image editing using GANs~\cite{patashnik2021styleclip, zhao2021focusgan, tao2023net}. FICE~\cite{pernuvs2023fice} introduced an innovative text-guided fashion editing model that combines CLIP semantic knowledge with GAN latent code optimization, guided by posture, regularization, and combination constraints. Diffusion-based models ~\cite{hertz2022prompt, couairon2023diffedit} have also demonstrated outstanding performance in general image editing tasks. Specifically, MGD~\cite{baldrati2023multimodal} was the first diffusion-based fashion image editing model, utilizing a U-Net denoising network that requires both text and sketch image as guiding conditions.

In summary, human-centric fashion editing methods are relatively lacking. Existing models either require complex inputs \textit{(e.g. keypoints, parsing images, masks, etc.)}, or are not designed for the fashion field. Furthermore, none of these models can effectively edit a wide range of apparel categories.

\begin{figure*}[t]
    \centering
    \includegraphics[width=\linewidth]{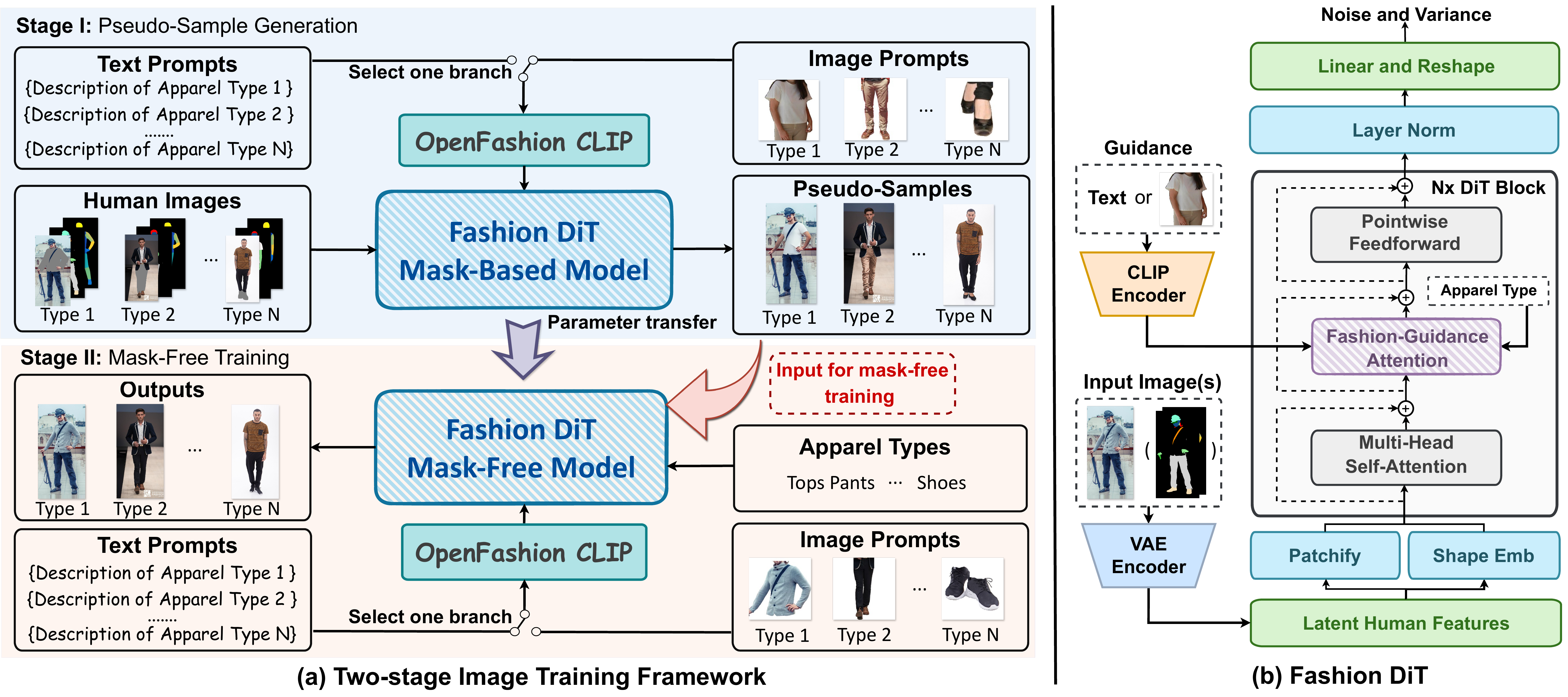}
    \caption{The Overall Architecture of Fashion Editing Framework. (a) Two-stage Image Training Framework. In Stage I, we train a mask-based model to generate pseudo-samples using unpaired text prompts or image prompts. In Stage II, we train the final mask-free utilizing generated pseudo-samples with the paired prompts and the apparel types as inputs. The training goal at this stage is to generate realistic images. (b) The architecture of Fashion DiT. }
    \label{fig:architecture}
\end{figure*}

\section{Method}
\label{sec:P3}
\textbf{Probelm Definition.} Specifically, given a person image \(P\in\mathbb{R}^{h \times w \times 3}\), this task requires synthesizing an edited image \(\hat{P} \in \mathbb{R}^{h \times w \times 3}\) based on either a textual prompt \(T\) or an image prompt \(G\) along with an apparel type label \(L\). The objective is to modify the targeted item while preserving the original appearance of the unedited regions in \(P\).

\subsection{Dataset Extension Method}
\label{sec:P3.1}
First, we introduce the method of extending the dataset. Existing fashion editing datasets such as VITON-HD~\cite{han2018viton} and Dresscode~\cite{morelli2022dress} focus on mainstream clothing items \textit{(i.e. tops, pants, and dresses)}, failing to meet extensive editing requirements like shoes and bags. SHHQ ~\cite{fu2022stylegan} offers human images with diverse apparel categories in complex backgrounds, but lacks critical information such as agnostic images and human densepose. 

To address this issue, we extend the SHHQ dataset to SHHQe, as shown in Fig~\ref{fig:agnostic_generation}. The extended dataset includes essential data: (1) 25 complete human keypoints extracted using openpose~\cite{cao2017realtime}, (2) densepose images obtained via~\cite{guler2018densepose}, (3) agnostic images and (4) guidance prompts obtained through dedicated extractors.

\subsubsection{Apparel-Agnostic Image Extraction.} Considering the differences among apparel, we apply category-specific mask strategies: strong removal for tops, pants, dresses, skirts, and shoes, and weak removal for scarves, bags, headwear, and socks.

•~\textit{Strong Removal}: Because the ``agnostic image'' created directly from parsing retains the contours of the original apparel, it limits the range of editable styles. The difference between apparel-only masking and wilder-area masking is shown in Fig~\ref{fig:agnostic_generation} (b). For apparel with large pieces, we apply the wilder-area masking to make the object shape indistinguishable. Specifically, for tops, we utilize the method from~\cite{choi2021viton} combined with keypoints, and we apply masks to areas such as the arms, torso, and neck. To prevent excessive removal, we use DensePose~\cite{guler2018densepose} combined with parsed labels to restore the position of hands and accessories. For pants, we expand the leg area towards the DensePose labels. For skirts or dresses, we generate a minimum bounding box for the lower garment area. For shoes, we integrate DensePose labels and parsed labels for wider area masking.

•~\textit{Weak Removal}: For more complex apparel items \textit{(i.e. scarves, bags, headwear, and socks)}, excessive removal of the apparel and surrounding areas could significantly compromise nearby details. Therefore, we employ a more cautious weak removal strategy, which involves creating masks only within the specific areas corresponding to the targeted apparel.

\subsubsection{Guidance Prompts Extraction.} Following data used in Openfasion CLIP~\cite{cartella2023openfashionclip}, we retained the related body regions to better preserve semantic information for tops, pants, dresses, shoes, and bags. For other apparel, masks from parse images are directly used to extract apparel regions. In the postprocessing, we deliberately place the apparel in the central position and set the background to pure white.

In this way, we extend the SHHQ to SHHQe, enhancing the dataset to support various apparel editing while containing complex backgrounds. For more details, please refer to the \textbf{supplementary material}.

\subsection{Mask-Free Image Editing Framework}
\label{sec:P3.2}

In this section, we introduce the method that enables users to edit any area using given instructions, depending on their preference. Furthermore, during inference, no auxiliary tools are required—only the mask-free diffusion model is necessary. As shown in Fig.~\(\ref{fig:architecture}\), the framework contains two stages.
\begin{itemize}
  \item \textbf{First stage:} Training a mask-based model to prepare high-quality pseudo-samples for the mask-free model. 
  \item \textbf{Second stage:} Constructing the final mask-free model which takes pseudo-samples as input. During the training, the original person images are used as the targets.
\end{itemize}
With the help of OpenFashion Clip~\cite{cartella2023openfashionclip}, both text prompts and image prompts can be supported. 

\subsubsection{Preliminary.} 
The diffusion models~\cite{Sohl2015deep,ho2020denoising} are a kind of generative models, which convert Gaussian noise into
samples via a reverse process of a fixed Markov Chain. We apply the denoising process on latent space using a pretrained autoencoder~\cite{esser2021taming}. This autoencoder compresses images \(x\in\mathbb{R}^{h \times w \times 3}\) into latent representations \(z=\mathcal{E}(x)\in\mathbb{R}^{ \frac{h}{f} \times \frac{w}{f} \times 4}\), where \(f\) represents the downsampling factor. We optimize the model with:
\begin{align} 
\mathcal{L}=E_{t, z_{t}, \epsilon\sim \mathcal N\left ( 0,1 \right )  }\left[\left\|\epsilon-\epsilon_{\theta}\left(z^t, t, conds\right)\right\|^{2}\right],
\end{align}
where \(t\) denotes the diffusion timestep, and \(conds\) are the selectable guiding conditions (\textit{e.g.} descriptive text prompt \(T\), image prompt \(G\) ). \(\epsilon_{\theta}\) represents the noise predicted by a DiT, which will be described in Sec.\ref{sec:P3.3}.

\subsubsection{Stage I: Pseudo-Sample Preparation with Mask-based model.} 
When training the mask-based diffusion model, the input comprises four components: (1) a paired text prompt \(T\) or image prompt \(G\); (2) an apparel-agnostic image \(P_m\in\mathbb{R}^{h \times w \times 3}\); (3) a human densepose image \(D\in\mathbb{R}^{h \times w \times 3}\); (4) a masked human parse image \(M\in\mathbb{R}^{h \times w \times 3}\), which is obtained as follows, 
\begin{align} 
M=(1-R_{agnostic})\odot{M_{raw}}.
\end{align}
\(R_{agnostic}\) represents the masked region in the agnostic image, and \(M_{raw}\) represents the original unprocessed human parse image. The symbol \(\odot\) represents the Hadamard product. The real person image \(\widetilde{P}\in\mathbb{R}^{h \times w \times 3}\) serves as the ground truth images.

During the inference, for each real person image \(\widetilde{P}\in\mathbb{R}^{h \times w \times 3}\), the mask-based model generates an edited version \(\hat{P}\in\mathbb{R}^{h \times w \times 3}\) using unpaired text \(T_u\) or image \(G_u\). Subsequently, the non-masked areas from the real image are overlaid with the generated image as: 
\begin{align} 
P=(1-R_{agnostic})\odot{\widetilde{P}}+R_{agnostic}\odot{\hat{P}}.
\end{align}
The \(P\), which is the final pseudo-sample will serve as the input for the next mask-free stage during training.

\subsubsection{Stage II: Mask-free Training.} 
Before training the final mask-free model, we first transfer the parameters from the shared structures of the mask-based model to the mask-free model. This strategy could leverage the knowledge from the mask-based model and shorten training time.

In the mask-free training stage, the model receives the reference pseudo-sample \(P\in\mathbb{R}^{h \times w \times 3}\), along with the paired textual description \(T\) or image prompt \(G\), as inputs. The real person image \(\widetilde{P}\in\mathbb{R}^{h \times w \times 3}\) serves as the ground truth image. This approach effectively preserves the poses in the reference image \(P\), even without an explicit mask, enabling the model to make precise and natural modifications to clothing or accessories. With a CLIP-based encoder, both the text and image prompts are supported.

\subsection{Fashion Diffusion Transformers (Fashion DiT)}
\label{sec:P3.3}
Recent studies reveal that transformers outperform U-Net-based methods concerning image denoising quality~\cite{bao2023all,peebles2023scalable} and efficiency~\cite{chen2023pixart}. As shown in Fig.~\ref{fig:architecture} (b), we propose Fashion DiT, an extension of DiT, as the denoising network backbone. Specifically, the images related to human figures are encoded as the latent space representations \(\mathcal{E}(P_m)\), \(\mathcal{E}(D)\), \(\mathcal{E}(M)\), and \(\mathcal{E}(\widetilde{P})\in\mathbb{R}^{ \frac{h}{f} \times \frac{w}{f} \times 4}\). These latent features are then divided into patches and processed through \(N \times\) DiT blocks. We incorporate shape embeddings from SDXL~\cite{podell2023sdxl} to handle various input resolutions. Using a text prompt as an example, the Fashion DiT Blocks can be expressed with a self-attention module \(\Phi^{\mathrm{A}}\), a feed-forward layer \(\Phi^{\mathrm{F}}\), and a Fashion-Guidance Attention (FGA) module as follows:

\begin{align}
\begin{split}
X_{i+1}=\Phi&^{\mathrm{F}}\circ\Phi^{\mathrm{FGA}}(I_{CLIP}, L, \Phi^{\mathrm{A}}(X_i))\\&I_{CLIP}=\mathcal{E}_{CLIP}(T),
\end{split}
\end{align}
where \(X_{i}\) denotes the input to the \(i\)-th DiT Block, \(L\) denotes the apparel type label. \(I_{CLIP}\) indicates the vector obtained by encoding the text or image prompt through the CLIP encoder. The output of the \(i\)-th DiT Block is \(X_{i+1}\).

\subsubsection{Fashion-Guidance Attention (FGA).} 
\begin{figure}[t]
    \centering
    \includegraphics[width=\linewidth]{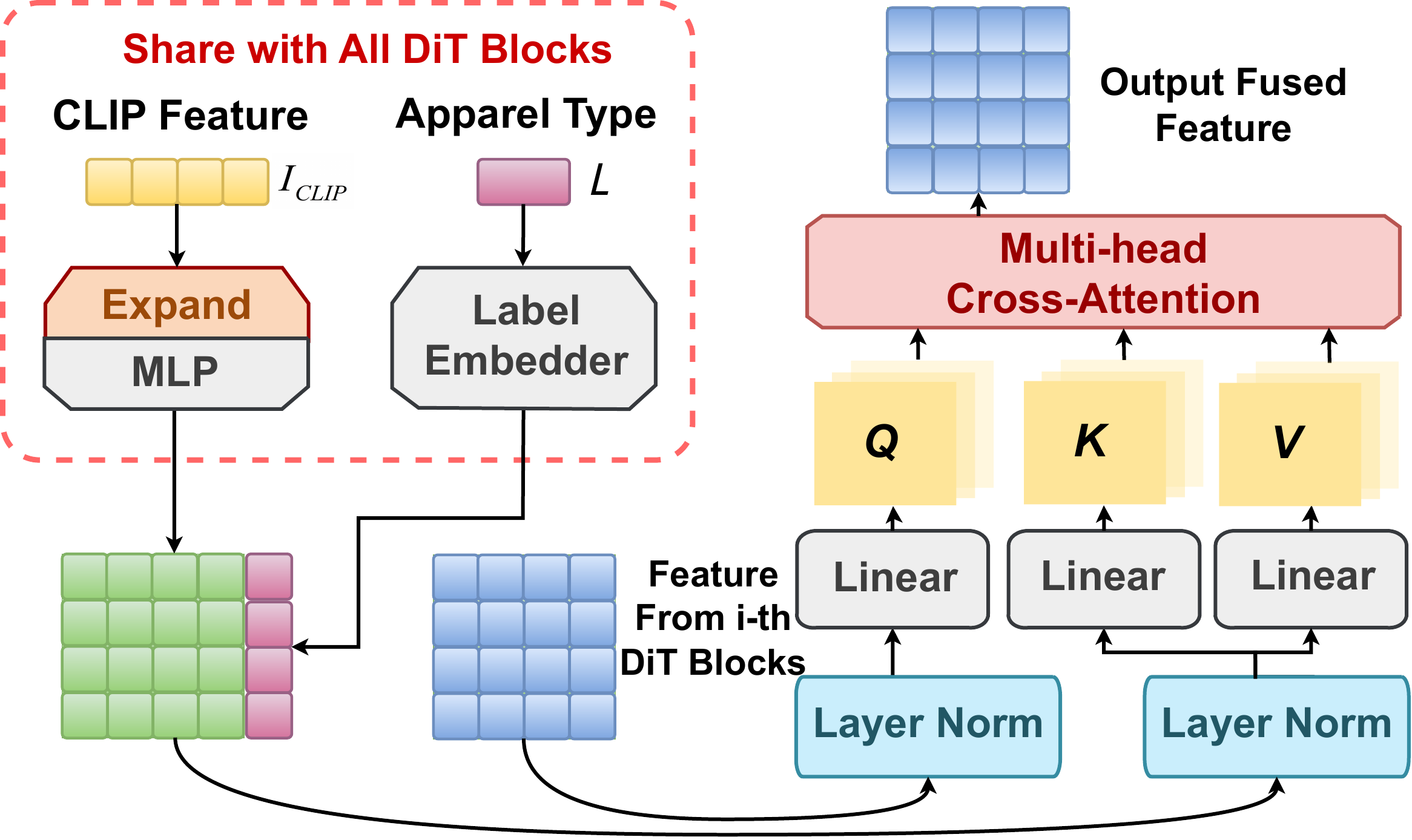}
    \caption{Fashion-Guidance Attention (FGA) Module.}
    \label{fig:feature_fusion}
\end{figure}

The Fashion-Guidance Attention (FGA) module is designed to integrate CLIP features with features from different apparel types, as illustrated in Fig.~\(\ref{fig:feature_fusion}\). Specifically, the apparel type \(L\) is first encoded into \(I_{L}\) via an embedding layer. \(I_{L}\) is first concatenated with \(I_{CLIP}\) after dimensionality expansion and Multi-Layer Perceptron (MLP) processing, resulting in a composite vector \(I_{F}\). After applying Layer Normalization (LN) to the \(I_{F}\), Key and Value in the attention mechanism are expressed as \(K = W_{k} \text{LN}(I_{F})\) and \(V=W_{v}\text{LN}(I_{F})\). The Queue is \(Q=W_{q}\text{LN}(I_{DiT})\), where \(I_{DiT}\) is the internal features of the previous transformer block. This fusion strategy aims to enable the model to thoroughly learn and focus on the regions corresponding to different apparel types, allowing for precise editing without the need for masked images. We also demonstrate the robustness and generalization capacity of the FGA module through the learned attention map, with further details in the \textbf{supplementary material}.

\begin{figure*}[t]
    \centering
    \includegraphics[width=1.0\linewidth]{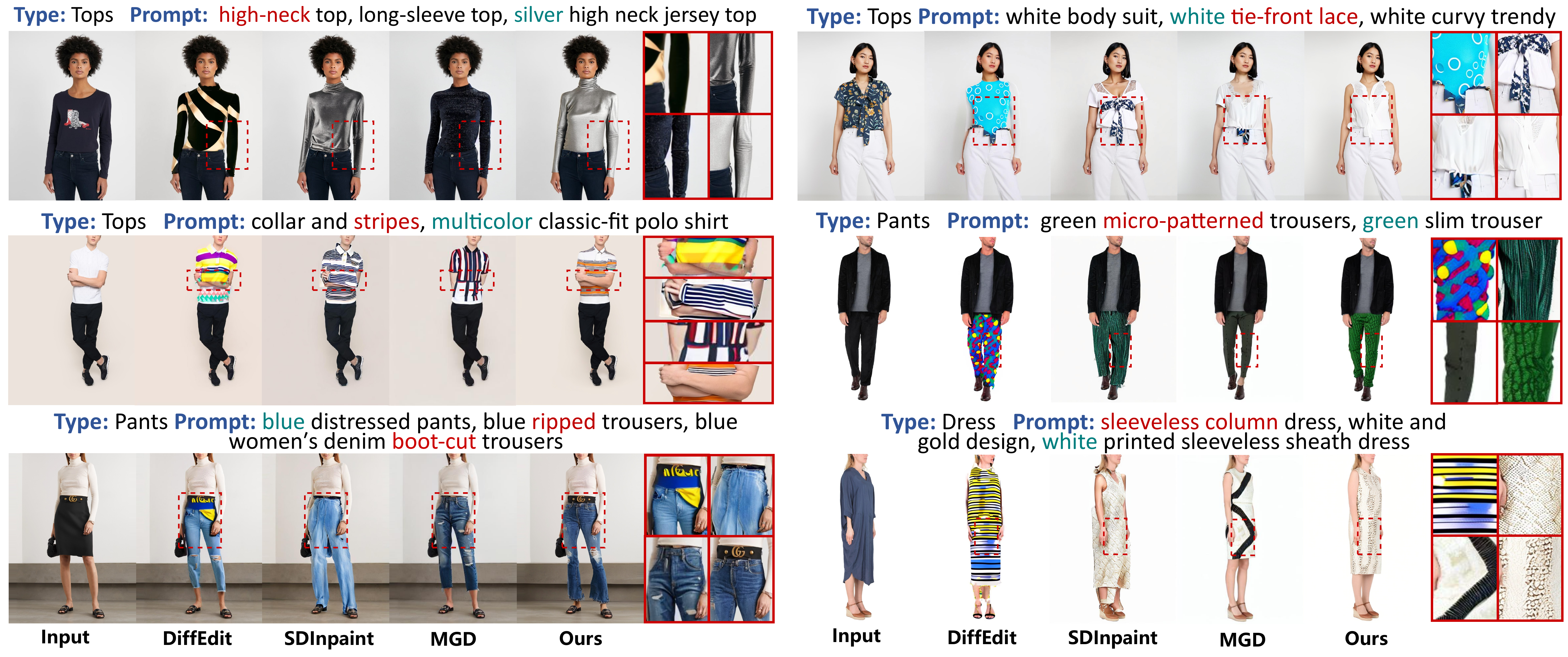}
    \caption{Visual Comparison on VITON-HD and Dresscode images. From left to right: the given person, the text-driven editing results by a series of methods.}
    \label{fig:compare_1}
\end{figure*}

\begin{figure*}[h!]
    \centering
    \includegraphics[width=\linewidth]{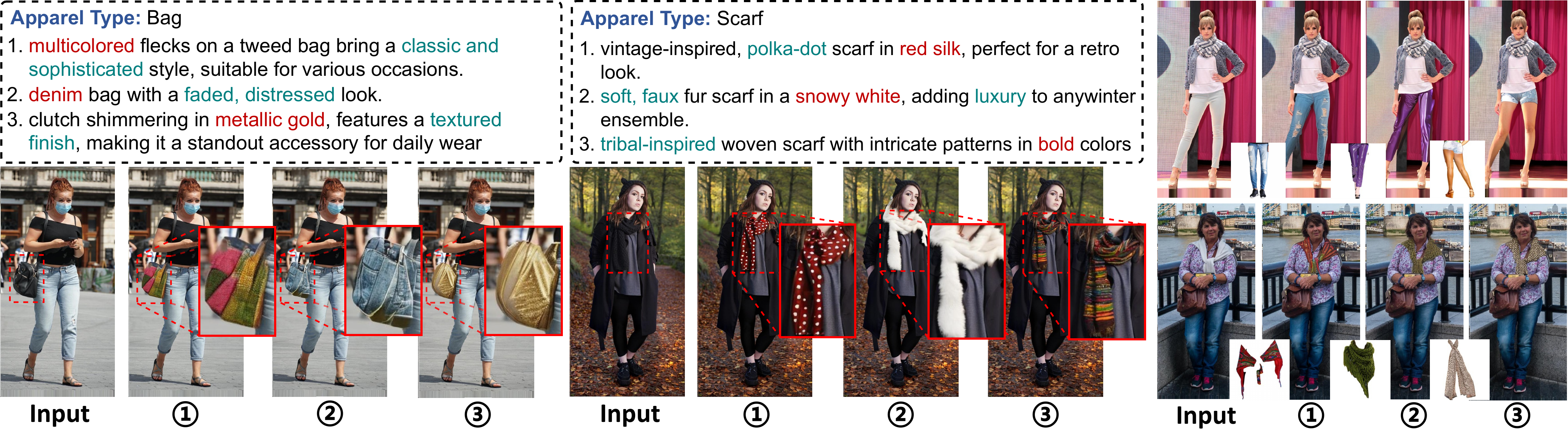}
    \caption{Visual results on SHHQe. In the text-driven editing, words indicating color are highlighted in red, while those indicating styles are highlighted in green. In image-driven editing, styles can be transferred. }
    \label{fig:in-the-wild}
\end{figure*}

\section{Experiments}
\label{sec:P4}

\subsection{Experimental Setup}
\label{sec:P4.1}
\subsubsection{Datasets.} We train our models on three datasets: VITON-HD~\cite{choi2021viton}, Dresscode~\cite{morelli2022dress}, and the constructed SHHQe. VITON-HD features image pairs of frontal-view women and upper body clothing. Dresscode contains upper body, lower body, and dress clothing image pairs. SHHQe contains a lot of raw human images, and encompasses a broad range of apparel items. For textual guidance in VITON-HD and Dresscode, we utilize text information provided by~\cite{baldrati2023multimodal}. We divide these datasets into training and test sets with 11,647/2,032 samples for VITON-HD and 48,392/5,400 for Dresscode. After extracting image prompts for each type of apparel in the SHHQe dataset, we obtained a total of 114,077 training and 12,653 testing samples including nine categories.

\subsubsection{Implementation Details.} In our setup, the scale factor \(f\) in the autoencoder is set to 8. 
For the denoising transformer, we configure the DiTs with a depth of 28 layers, set the channel size to 1,152, the patch size to 2, and the number of heads in the cross-attention layers to 16. We use the Adam optimizer, with a learning rate of \(1e^{-4}\) and 1,000 time steps. During the inference stage, the model employs SA-solver~\cite{xue2023stochastic} for sampling, with a classifier-free guidance scale \(s\) of 4.5. All experiments are conducted on NVIDIA Tesla H800 GPUs.

\begin{table*}[t]

    \centering
    \caption{Quantitative results on the VITON-HD and Dresscode datasets. The KID is scaled by 1000 for better comparison. The best and second best are marked in bold
and underlined, respectively.}
    \label{tab:fid}\tiny
    \setlength{\tabcolsep}{.3em}
    \resizebox{\linewidth}{!}{
    \begin{tabular}{l c ccccc c ccccc }
    \toprule

     \multirow{2}{*}{\textbf{Method}} & \multirow{2}{*}{\makecell{\textbf{Mask-free}}} & 
     \multicolumn{5}{c}{\textbf{VITON-HD}} & &  \multicolumn{5}{c}{\textbf{Dresscode}} \\
     \cmidrule{3-7} \cmidrule{9-13} 
     
     & & \textbf{FID}\(\downarrow\) & \textbf{KID}\(\downarrow\) & \textbf{CLIP-S}\(\uparrow\) & \textbf{LPIPS$_\text{p}$}\(\downarrow\) & \textbf{SSIM$_\text{p}$}\(\uparrow\) & & \textbf{FID}\(\downarrow\) & \textbf{KID}\(\downarrow\) & \textbf{CLIP-S}\(\uparrow\) & \textbf{LPIPS$_\text{p}$}\(\downarrow\) & \textbf{SSIM$_\text{p}$}\(\uparrow\) \\
    
    \midrule
    \textbf{256$\times$192 resolution} \\
    FICE~\cite{pernuvs2023fice} & \ding{55} & 52.74 & 48.58 & 25.94 & - & - &  & 34.14 & 26.86 & 26.03 & - & - \\ 
    DiffEdit~\cite{couairon2023diffedit} & \checkmark & 36.13 & 15.25 & 23.53 & 0.232 & 0.738 &  & 19.02 & 6.21 & 20.71 & 0.195 & 0.764\\ 
    SDInpaint~\cite{runwayml_stable_diffusion_inpainting} & \ding{55} & \underline{11.47} & 3.19 & \underline{30.16} & 0.147 & 0.828 & & 13.06 & 5.91 & 28.13 & 0.133 & 0.860  \\ 
    MGD~\cite{baldrati2023multimodal}  & \ding{55} & 11.54 & \underline{3.18} & 29.95 & \underline{0.145} & \underline{0.838} & &  \underline{7.01} & \underline{2.19} & \underline{29.58} & \textbf{0.119} & \textbf{0.880} \\ 
    \rowcolor{LightCyan}
    \textbf{Ours} & \checkmark & \textbf{8.56} & \textbf{0.69} & \textbf{30.54} & \textbf{0.126} & \textbf{0.843} & & \textbf{6.09} & \textbf{0.79} & \textbf{30.66} & \underline{0.123} & \underline{0.875} \\ 
    
    \midrule
    \textbf{512$\times$384 resolution} \\
    SDedit~\cite{meng2022sdedit} & \ding{55}  & 15.12 & 5.67 & 28.61  & - & - & &  11.38 & 5.69 & 27.10 & - & - \\ 
    DiffEdit~\cite{couairon2023diffedit} & \checkmark & 28.82 & 17.00 & 23.80 & 0.257 & 0.733 &  & 21.06 & 6.91 & 21.85 & 0.202 & 0.759 \\ 
    SDInpaint~\cite{runwayml_stable_diffusion_inpainting} & \ding{55} & \underline{12.37} & \underline{3.48} & 30.13 & \underline{0.155} & 0.824 & & 14.09 & 6.51 & 29.05 &  
 0.138 & 0.860  \\ 
    MGD~\cite{baldrati2023multimodal}  & \ding{55} & 12.80 & 3.86 & \underline{30.75} & 0.156 & \underline{0.832} &  &  \underline{7.73} & \underline{2.82} & \underline{30.04} & \underline{0.127} & \underline{0.871} \\ 
    \rowcolor{LightCyan}
    \textbf{Ours} & \checkmark & \textbf{9.16} & \textbf{0.68} & \textbf{31.25} &  \textbf{0.139} & \textbf{0.839} &  & \textbf{6.56} & \textbf{0.91} & \textbf{31.24} & \textbf{0.124} & \textbf{0.873} \\ 

    \midrule
    \textbf{1024$\times$768 resolution} \\
    DiffEdit~\cite{couairon2023diffedit} & \checkmark & 30.26 & 18.06 & 24.03 & 0.276 & 0.727 &  & 22.03 & 7.65 & 22.53 & 0.227 & 0.762 \\ 
    SDInpaint~\cite{runwayml_stable_diffusion_inpainting} & \ding{55} & \underline{12.99} & \underline{3.89} & \underline{30.89} & \underline{0.174} & \underline{0.828} & & \underline{15.16} & \underline{7.33} & \underline{29.96} & \underline{0.155} & \underline{0.852}   \\ 
    \rowcolor{LightCyan}
    \rowcolor{LightCyan}
    \textbf{Ours} & \checkmark & \textbf{9.43} & \textbf{0.57} & \textbf{31.74} & \textbf{0.151} & \textbf{0.836} &  & \textbf{6.72} &  \textbf{0.84} & \textbf{31.91} & \textbf{0.130} & \textbf{0.866} \\ 
       
    \bottomrule
    \end{tabular}
    }
\vspace{-.2cm}
\end{table*}

\subsection{Visual Results}
\label{sec:P4.2}

\subsubsection{Comparation with Existing Methods.} In Fig.~\(\ref{fig:compare_1}\), We compare our model with the competitive text-driven fashion editing models DiffEdit~\cite{couairon2023diffedit}, SDInpaint~\cite{runwayml_stable_diffusion_inpainting} and MGD~\cite{baldrati2023multimodal}.
For fair comparison, experiments are conducted on VITON-HD and Dresscode datasets. Our model obviously outperforms others with respect to semantic consistency \textit{(e.g \textbf{``high-neck silver top"}, \textbf{``green micro-patterned trousers"}, \textbf{``white and gold sleeveless column dress"})} and image synthesis quality \textit{(e.g \textbf{``white tie-front lace"},  \textbf{``multicolor striped polo"})}.

\subsubsection{Visual Results on SHHQe dataset.} 
In Fig.~\(\ref{fig:in-the-wild}\),
visual results on the SHHQe dataset are presented. Both the text-driven and the style-image-driven editing results are promising. For instance, our model can change the appearance of \textbf{\textit{a woman's scarf}} from \textbf{\textit{black}} to \textbf{\textit{snowy white luxury}} with text instruction, and can change the \textbf{\textit{pants}} into \textbf{\textit{shorts}} with an apparel image. More visual examples are in \textbf{supplementary material}.

\subsection{Quantitative Comparison}
\label{sec:P4.3}
Following~\cite{baldrati2023multimodal},
we use Fréchet Inception Distance (FID)~\cite{heusel2017gans} and Kernel Inception Distance (KID)~\cite{sutherland2018demystifying} to measure the distributed distance between the generated and expected images, and we employ the CLIP Score~\cite{hessel2021clipscore} to evaluate the consistency between the generated contents and text descriptions. The above metrics are all evaluated using the unpaired setting. Then, we use Structural Similarity (SSIM)~\cite{wang2004image} and Learned Perceptual Image Patch Similarity (LPIPS)~\cite{zhang2018unreasonable} to measure the similarity between the generated images and expected ones, where the paired setting is adopted. For fairness, we perform a detailed comparison using text prompts at various resolutions on the VITON-HD and Dresscode datasets. The results are shown in Table~\(\ref{tab:fid}\). Our model significantly outperforms the mask-free editing method DiffEdit~\cite{couairon2023diffedit} across all metrics. Additionally, our method outperforms mask-based approaches across most metrics.

\subsection{Human Evaluation} 
\label{sec:P4.4}
\begin{figure}[h]
    \centering
    \includegraphics[width=\linewidth]{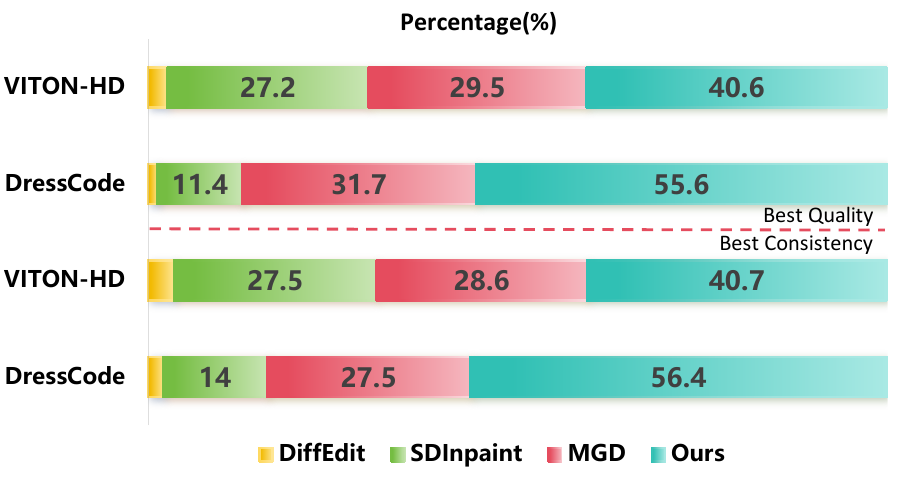}
    \caption{The human evaluation results in terms of image quality and semantic consistency. }
    \label{fig:human_eval}
\end{figure}

To evaluate our model in terms of human perception, we present the user study on 50 human participants. We randomly select 40 sets of data, each containing results from different methods. Each participant is asked to choose the best samples in terms of quality and semantic consistency. In Figure~\ref{fig:human_eval}, the \textit{Jab} scores~\cite{bhunia2023person} are shown, which represents the percentage of images considered the best among all methods. It can be seen that our model demonstrates outstanding performance against three baselines on two datasets. 

\subsection{Ablation Study} 
\label{sec:P4.5}

\subsubsection{Ablation Study on  VITON-HD and Dresscode.} Firstly, ablation experiments are conducted on public datasets VITON-HD and Dresscode, as shown in Table~\(\ref{tab:ablition}\). We present the results from the mask-based model (M.B.) and the mask-free model (M.F.) with or without FGA. It can be observed that M.F. outperforms M.B., despite the fact that M.F. was trained on data generated by the M.B. model. This superior performance can be attributed to the transfer of parameters from M.B. to M.F., allowing M.F. to leverage the knowledge learned by M.B. The noise in the pseudo-samples also contributes to the enhanced robustness of M.F. Furthermore, the model exhibited additional performance improvements after incorporating the FGA.

\begin{table}[h]
    \centering
    \caption{Ablation study on VITON-HD and Dresscode datasets at resolution $1024 \times 768$ using text guidance.}
    \label{tab:ablition}
    \setlength{\tabcolsep}{.3em}
    \resizebox{\linewidth}{!}{ 
    \begin{tabular}{l ccc c ccc }
    \toprule
    \multirow{2}{*}{\textbf{Model Type}} & \multicolumn{3}{c}{\textbf{VITON-HD}} & & \multicolumn{3}{c}{\textbf{Dresscode}} \\
    \cmidrule{2-4} \cmidrule{6-8}
     & \textbf{FID}$\downarrow$ & \textbf{KID}$\downarrow$ & \textbf{CLIP-S}$\uparrow$ & &  \textbf{FID}$\downarrow$ & \textbf{KID}$\downarrow$ & \textbf{CLIP-S}$\uparrow$ \\
    \midrule
    M.B. & 10.34 & 1.21 & 31.34 & & 7.42 & 1.91 & 31.07 \\
    M.F. w/o FGA & 9.70 & 0.85 & 31.52 & & 7.21 & 1.43 & 30.18  \\
    \rowcolor{LightCyan}
    \textbf{M.F. w/i FGA (Full Model)}
    & \textbf{9.43} & \textbf{0.57} & \textbf{31.74} & & \textbf{6.72} & \textbf{0.84} & \textbf{31.91} \\
    \bottomrule
    \end{tabular}
    }
\end{table}

\begin{table}[h]
    \centering
    \caption{Ablation study on SHHQe using text and image guidance. CS: CLIP Text score. CIS: CLIP Image score.}
    \label{tab:ablition_shhq}
    \setlength{\tabcolsep}{.3em}
    \resizebox{\linewidth}{!}{ 
    \begin{tabular}{l ccc c ccc }
    \toprule
    \multirow{2}{*}{\textbf{Model Type}} & \multicolumn{3}{c}{\textbf{Text-Driven}} & & \multicolumn{3}{c}{\textbf{Image-Driven}} \\
    \cmidrule{2-4} \cmidrule{6-8}
     & \textbf{FID}$\downarrow$ & \textbf{KID}$\downarrow$ & \textbf{CS}$\uparrow$ & &  \textbf{FID}$\downarrow$ & \textbf{KID}$\downarrow$ & \textbf{CIS}$\uparrow$ \\
    \midrule
    M.B. & 4.80 & 0.79 & 23.09 & & \textbf{4.05} & 0.47 & 56.01 \\
    M.F. w/o FGA & 4.97 & 0.72 & 18.42 & & 4.73 & 0.58 & 55.76  \\
    \rowcolor{LightCyan}
    \textbf{M.F. w/i FGA (Full Model)} & \textbf{4.67} & \textbf{0.65} & \textbf{24.02} & & 4.26 & \textbf{0.45} & \textbf{58.22} \\
    \bottomrule
    \end{tabular}
    }
\end{table}

\subsubsection{Ablation Study on SHHQe Dataset.} 
We conduct a comprehensive quantitative ablative evaluation of the SHHQe dataset. Both the text-driven and the image-driven ablation results are presented in Table~\(\ref{tab:ablition_shhq}\) and Table~\(\ref{tab:ablition_shhq_ca}\). In the latter, smaller apparel items such as bags, scarves, headwear, and socks are grouped together under the category of accessories. The leading metrics across various models further highlight the overall superiority of the full model. For more evaluation in the category of accessories, please refer to the \textbf{supplementary material}. 

\begin{table}[h!]
    \centering
    \caption{Category-wise CLIP Score on SHHQe.}
    \label{tab:ablition_shhq_ca}
    \setlength{\tabcolsep}{.3em}
    \resizebox{\linewidth}{!}{
    \begin{tabular}{lccc}
    \toprule
    

    \textbf{Model Type} & \textbf{Category} & \textbf{CS}\(\uparrow\)(Text-Driven) & \textbf{CIS}\(\uparrow\)(Image-Driven) \\
    \midrule

    \multirow{5}{*}{M.B.} & Tops & 26.04 & 50.24 \\
    & Pants & 26.06 & 57.77 \\
    & Dresses\&Skirts & 25.24 & 55.04 \\
    & Shoes & 20.52 & 61.19 \\
    & Accessories  & 17.81 & \textbf{52.87} \\
    \midrule
    \multirow{5}{*}{M.F. w/o FGA} & Tops & 25.83 & \textbf{54.85} \\
    & Pants & 15.98 & 53.31 \\
    & Dresses\&Skirts & 20.90 & 55.98 \\
    & Shoes & 14.33 & 60.21 \\
    & Accessories  & 15.87 & 50.58 \\
    \midrule
    \multirow{5}{*}{\makecell{\textbf{M.F. w/i FGA} \\ \textbf{(Full Model)}}}
    & Tops & \textbf{27.82} & 54.71 \\
    & Pants & \textbf{26.93} & \textbf{59.67} \\
    & Dresses\&Skirts & \textbf{27.18} & \textbf{58.53} \\
    & Shoes & \textbf{20.76} & \textbf{62.64} \\
    & Accessories  & \textbf{17.86} & 52.12 \\

    \bottomrule
    \end{tabular}
    }
\end{table}

\vspace{-0.2cm}
\section{Conclution}
In this paper, we introduce a data extension method and propose \textbf{AnyDesign}, a diffusion-based mask-free framework for human fashion editing. The proposed model can edit in-the-wild human images guided by texts or style images. In the framework, an innovative Fashion DiT with the Fashion-Guidance Attention module is further introduced to enhance the flexibility of editing versatile types of apparel. Experiments demonstrate that our method performs excellently in human fashion editing and achieves state-of-the-art performance in popular datasets. We believe our work will advance fashion editing technology for real-world use.
\newpage

\bibliography{aaai25}
\appendix
\twocolumn[
\begin{center}
    \LARGE
    \vspace{1em} 
    \textbf{AnyDesign: Versatile Area Fashion Editing via Mask-Free Diffusion\\--~Supplementary Material~--}
\end{center}
\vspace{1em} 
]

\textbf{Overview:} In this supplementary material, we first provide more experimental details including visual and quantitative results, the visual attention map learned from the FGA module, and the classifier-free diffusion guidance technique used for sampling. Next, we provide further details on the SHHQe dataset, including statistical analyses and examples of the processed samples. Lastly, we delve into a discussion of our work, encompassing its limitations as well as potential avenues for future research.

\section{More Experimental Results}\label{secA3}
\subsection{More Visual Results}
\begin{figure*}[h!]
    \centering
    \includegraphics[width=0.78\linewidth]{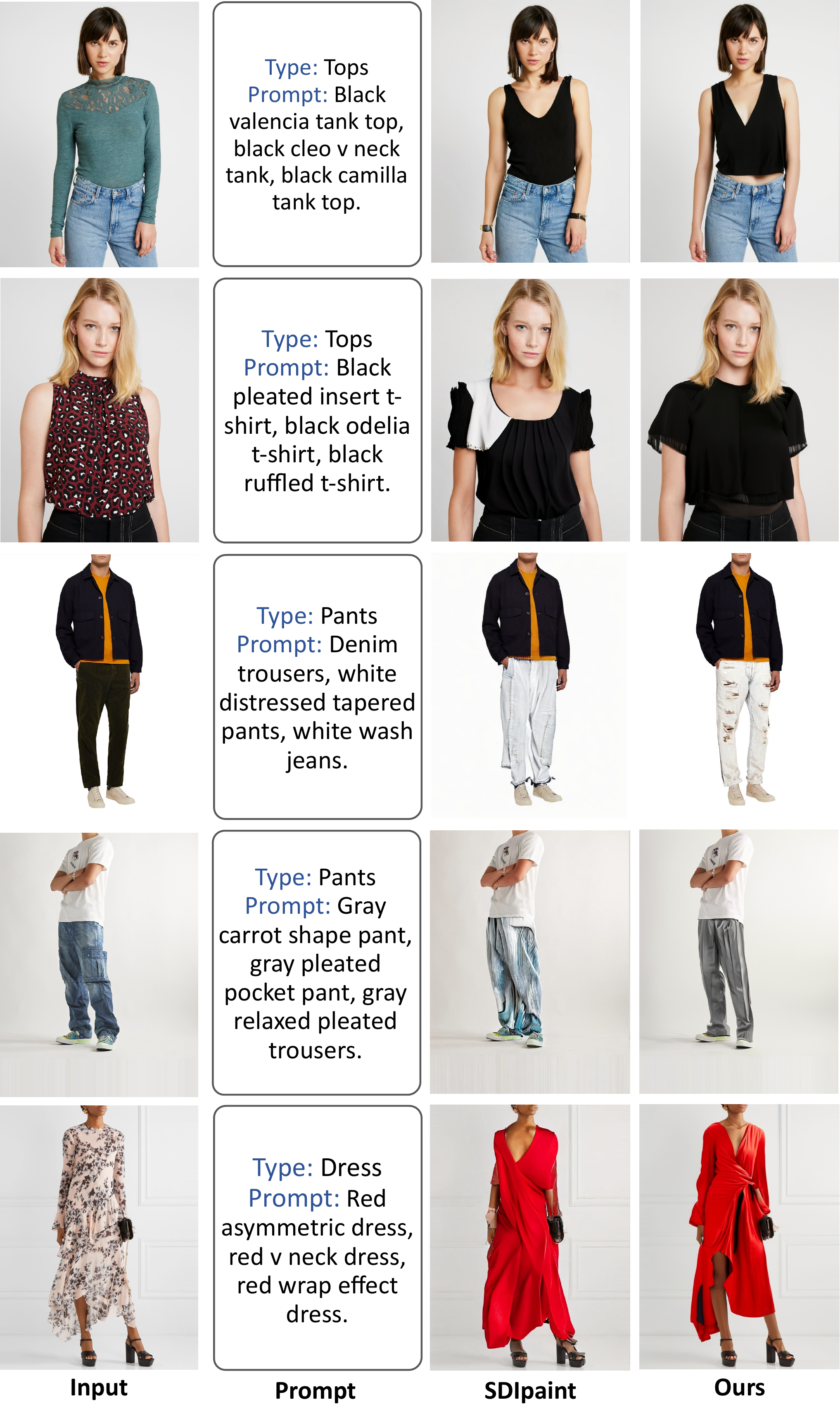}
    \caption{Visual Comparison of our model and SDInpaint.}
    \label{fig:compare_sdinpaint}
\end{figure*}

\begin{figure*}[h!]
    \centering
    \includegraphics[width=0.78\linewidth]{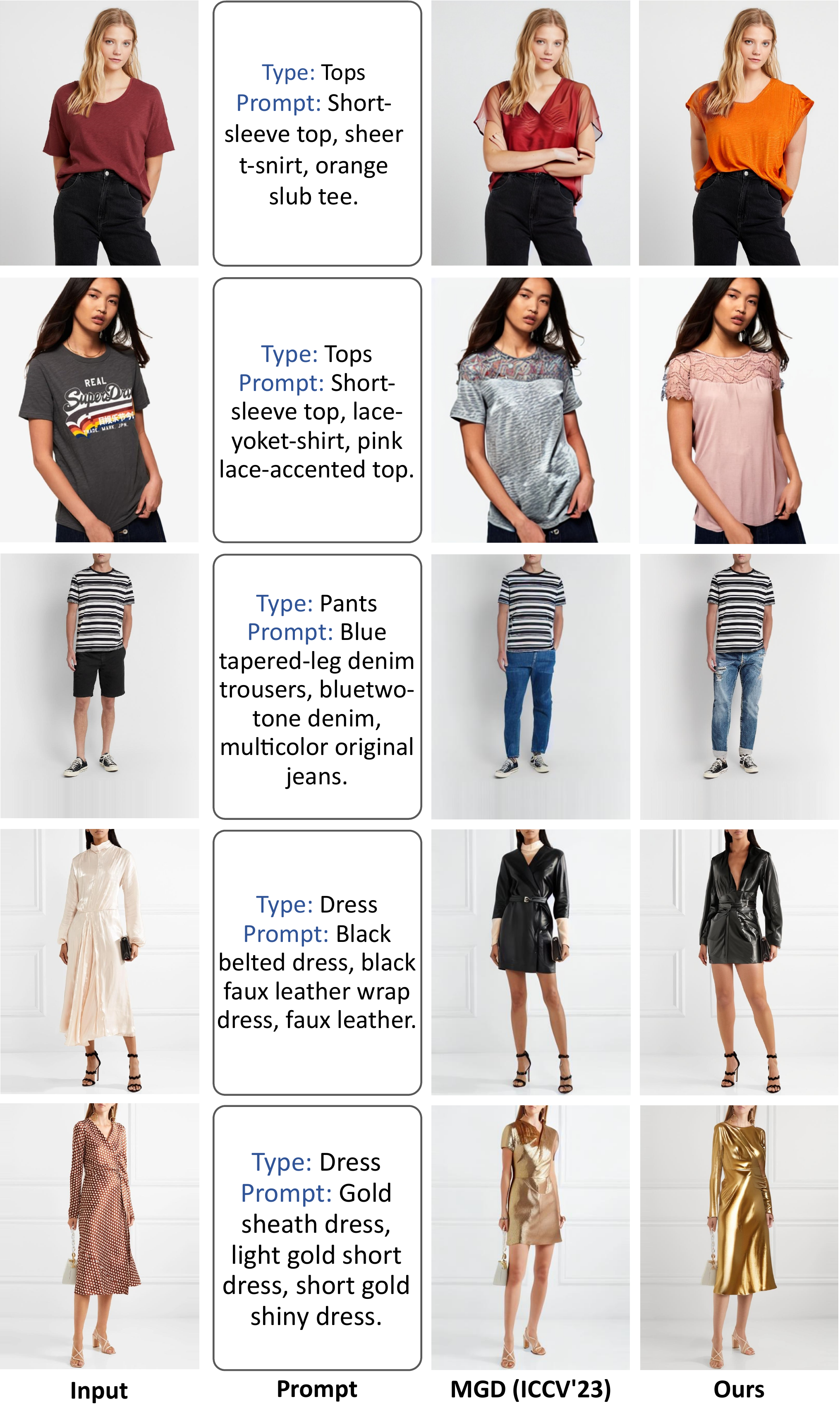}
    \caption{Visual Comparison of our model and MGD.}
    \label{fig:compare_mgd}
\end{figure*}

\begin{figure*}[h!]
    \centering
    \includegraphics[width=0.72\linewidth]{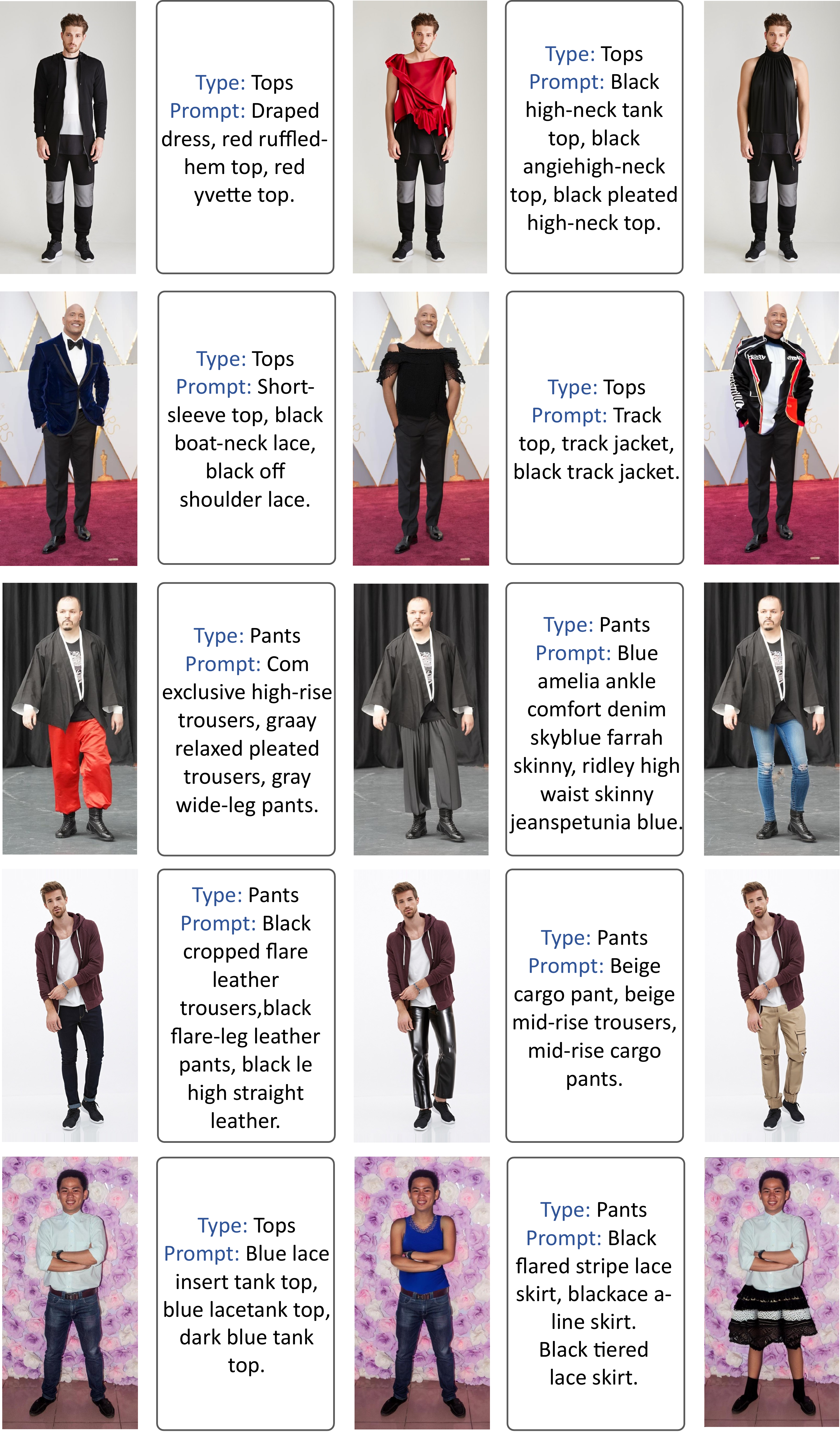}
    \caption{Editing results of various apparel on SHHQe.}
    \label{fig:shhq_1}
\end{figure*}

\begin{figure*}[h!]
    \centering
    \includegraphics[width=0.72\linewidth]{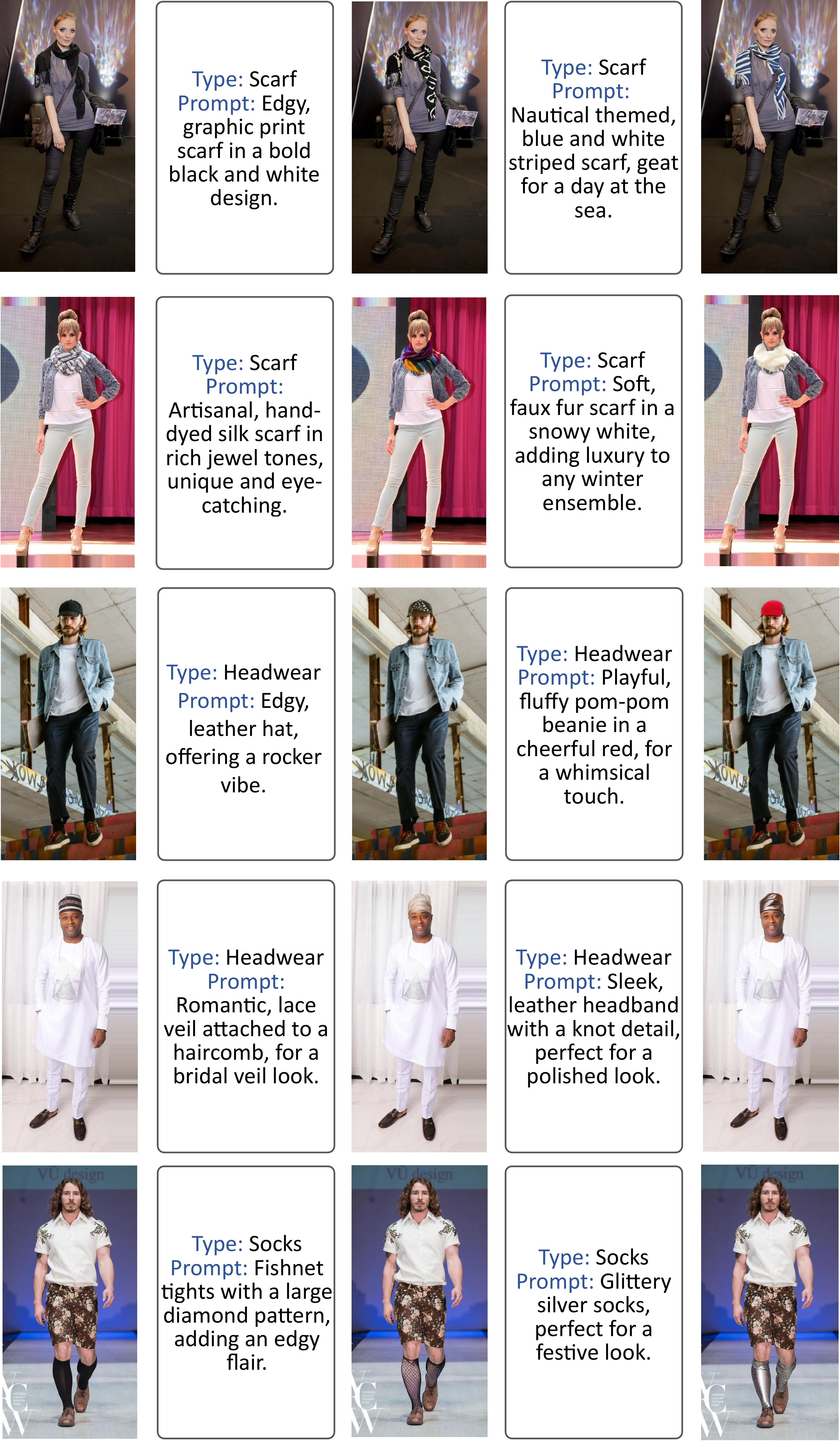}
    \caption{Editing results of various apparel on SHHQe.}
    \label{fig:shhq_2}
\end{figure*}

\begin{figure*}[h!]
    \centering
    \includegraphics[width=0.72\linewidth]{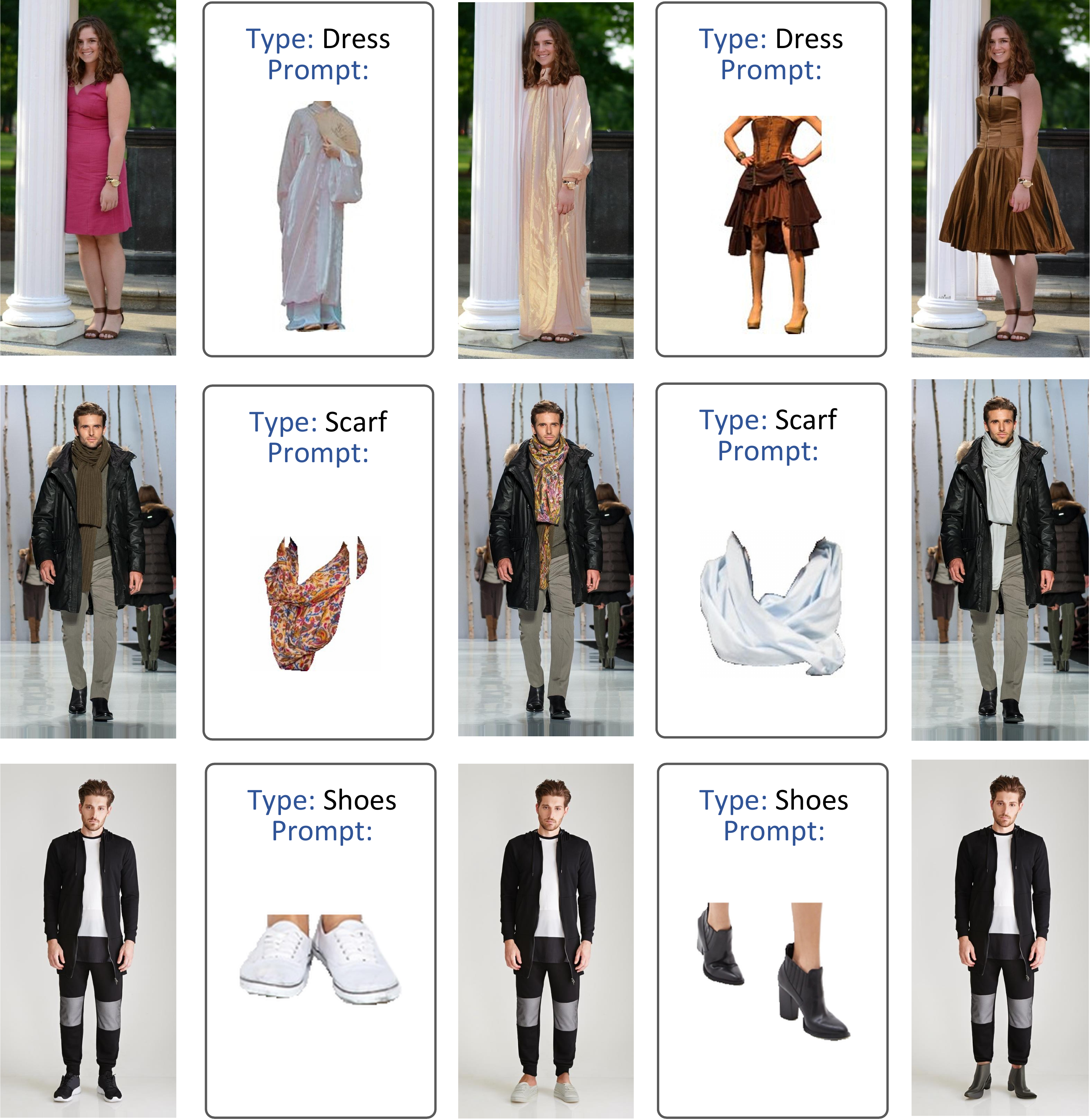}
    \caption{Editing results of various apparel on SHHQe.}
    \label{fig:shhq_3}
\end{figure*}

Figures~\(\ref{fig:compare_sdinpaint}\), and~\(\ref{fig:compare_mgd}\) show the visual comparison of three competitive methods. Figures~\(\ref{fig:shhq_1}\), \(\ref{fig:shhq_2}\), and \(\ref{fig:shhq_3}\) present additional examples of text-driven and style-image-driven results across various apparel categories. These examples further showcase the model's advanced fashion editing features and visual performance.

\subsection{Additional Ablation Results}

\noindent\textbf{More Ablation Results on SHHQe.} Tab. \(\ref{tab:shhq}\) presents detailed evaluation results in terms of CLIP Score across \textit{dress}, \textit{skirt} and\textit{ all accessory categories}, driven by text prompts and image prompts, further demonstrating the significant advantage of our model.

\begin{table}[h!]
    \centering
    \caption{More results in terms of CLIP Score on SHHQe. CS: CLIP Text Score. CIS: CLIP Image Score.}
    \label{tab:shhq}
    \setlength{\tabcolsep}{.3em}
    \resizebox{\linewidth}{!}{
    \begin{tabular}{lccc}
    \toprule
    
    \multirow{2}{*}{\textbf{Model Type}} & \multirow{2}{*}{\textbf{Category}} & \multirow{2}{*}{\makecell{\textbf{CS}\(\uparrow\) \\ (Text-Driven)}} & \multirow{2}{*}{\makecell{\textbf{CIS}\(\uparrow\) \\ (Image-Driven)}} \\
    \\
    \midrule

    \multirow{5}{*}{M.B.} & Dress & 24.79 & 54.32  \\
    & Skirt & 26.39  & 56.89  \\
    & Scarf &  22.35  & 56.44  \\
    & Socks &  17.63  & 53.03  \\
    & Headwear  &  15.75  & 53.42  \\
    & Bag  &  19.27  & 51.47  \\
    \midrule
    \multirow{5}{*}{M.F. w/o FGA} & Dress &  21.83 & 56.28  \\
    & Skirt & 18.49 & 55.18  \\
    & Scarf & 20.63 & 54.25  \\
    & Socks & 14.95 & 49.55  \\
    & Headwear  & 13.68 & 52.55 \\
    & Bag  & 15.71 & 48.68  \\
    \midrule
    \multirow{5}{*}{\makecell{\textbf{M.F. w/i FGA} \\ \textbf{(Full Model)}}}
    & Dress & \textbf{26.94} & \textbf{58.10} \\
    & Skirt & \textbf{27.79} & \textbf{59.65} \\
    & Scarf & \textbf{23.35} & 56.22 \\
    & Socks & \textbf{18.59} & 51.10 \\
    & Headwear  & \textbf{15.90} & \textbf{53.71} \\
    & Bag  & 18.26 & 50.53 \\
    
    \bottomrule
    \end{tabular}
    }
\end{table}

\begin{figure}[h]
    \centering
    \includegraphics[width=\linewidth]{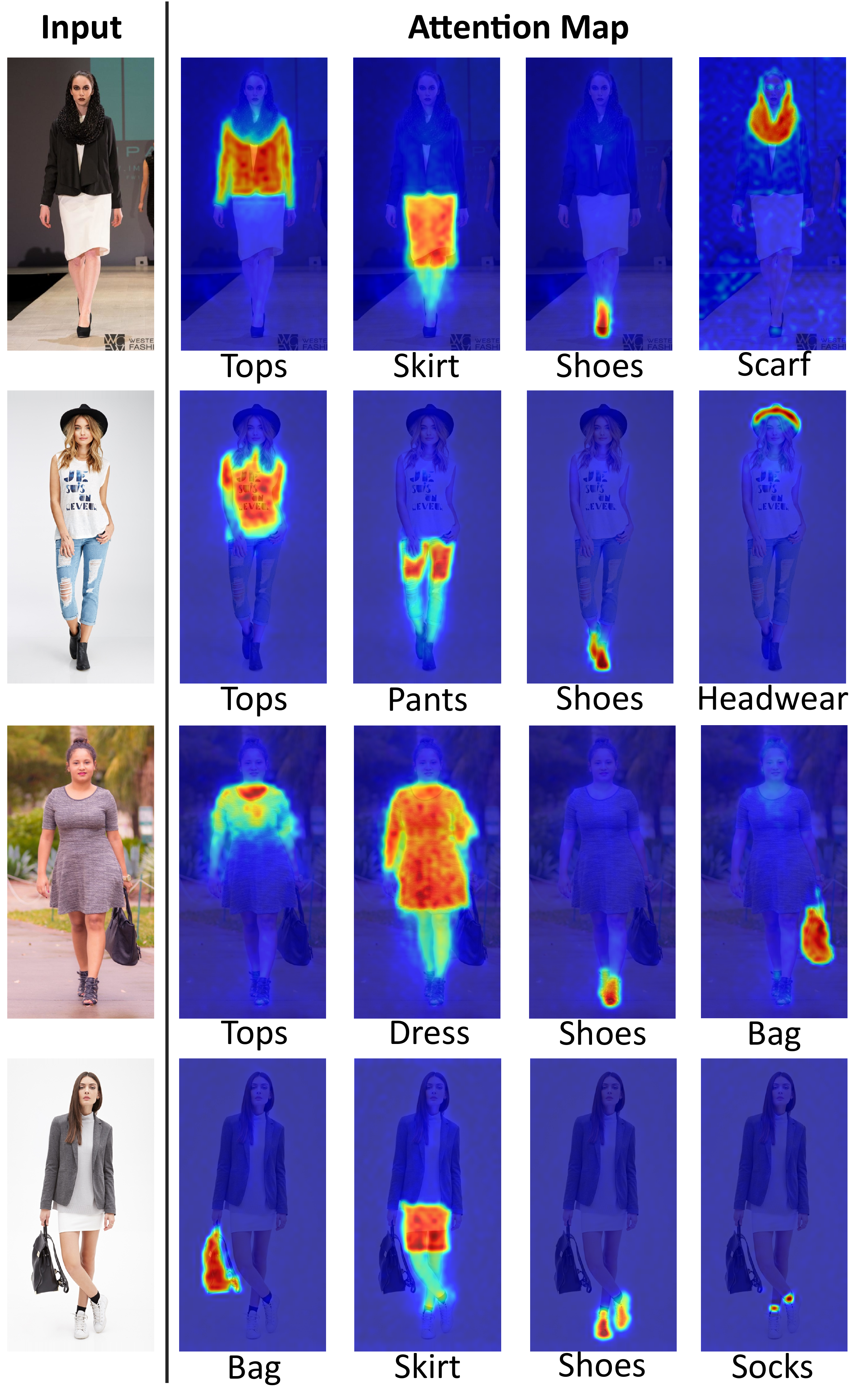}
    \caption{Visualization of attention maps learned by FGA module, using different apparel types as guidance.}
    \label{fig:attention_map}
\end{figure}

\noindent\textbf{Attention Map Learned from FGA module.} We use captions that do not contain specific apparel information, combined with different type labels as conditions, to virtualize attention maps learned by FGA module (Fig.~\(\ref{fig:attention_map}\)). This figure illustrates how the fashion guidance module aids the model in accurately identifying the locations of various apparel items, thereby enabling precise editing.

\begin{table*}[t]

  \caption{Text description examples generated by GPT-4.
  }
  \label{tab:description}
  \centering
  \begin{tabular}{@{}p{1.5cm}<{\centering}p{15.5cm}@{}}
    \toprule
    \multicolumn{1}{c}{\textbf{Category}} & \multicolumn{1}{c}{\textbf{Textual Descriptions}}  \\
    \midrule
    \multirow{2}{*}{Tops}  &  1. Vintage-inspired, velvet top with puff sleeves, offering a touch of retro glamour. \newline  2. Trendy, balloon sleeve blouse in emerald green, adding a dramatic flair to skinny jeans.  \\
    \midrule
    \multirow{2}{*}{Pants}  & 1. Classic, straight-leg trousers in navy blue, a wardrobe staple for versatility. \newline 2. Sophisticated, wide-leg pants in a timeless black and white plaid, ideal for the office or evenings out.  \\
    \midrule
    \multirow{2}{*}{Dress}  & 1. Trendy, ruched bodycon dress with a vibrant tie-dye print, for a modern and colorful look. \newline 2. Chic, bohemian maxi dress with a tribal-inspired pattern, perfect for a beachy style.  \\
    \midrule
    \multirow{2}{*}{Skirt}  & 1. Casual, button-front denim skirt with a distressed finish, for a laid-back and trendy look. \newline 2. Edgy, faux leather midi skirt with a slit detail, adding an alternative touch.   \\
    \midrule
    \multirow{2}{*}{Shoes}  &  1. Vintage, wingtip brogue shoes in two-tone brown and white, adding retro charm. \newline 2. Playful, platform sneakers in a bold leopard print, for a trendy and casual look. \\
    \midrule
    \multirow{2}{*}{Scarf}  & 1. Soft, faux fur scarf in a snowy white, adding luxury to any winter ensemble. \newline 2. Vintage-inspired, polka-dot scarf in red silk, perfect for a retro look.   \\
    \midrule
    \multirow{2}{*}{Bag}  &  1. Rugged, waxed canvas field bag in olive drab, with leather buckles and an adjustable shoulder strap. \newline 2. Minimalist, felt bag sleeve in charcoal, with a secure snap button closure.   \\
    \midrule
    \multirow{2}{*}{Headwear}  &  1. Preppy, straw boater headwear with a stripe ribbon, for a nau \newline 2. Sleek, leather headband with a knot detail, perfect for a polished look.  \\
    \midrule
    \multirow{2}{*}{Socks}  &  1. Cozy, thick-knit wool socks in a classic grey, perfect for chilly winter days.  \newline 2. Bright, neon green athletic socks made from moisture-wicking fabric for high-performance activities. \\
  \bottomrule
  \end{tabular}
\end{table*}

\subsection{Guidance Technique for Sampling.} Following the strategy of classifier-free diffusion guidance~\cite{ho2022classifier}, we jointly train the model by randomly emptying the conditions. In the sampling process, the model performs sampling through a linear combination of conditional and unconditional estimated noises: 
\begin{align}
\begin{split}
\hat{\epsilon}_{\theta}(z_{\hat{P}}^t|I_{CLIP}, L)&=\epsilon_{\theta}(Z_{\hat{P}}^t|I_\emptyset,L_\emptyset)\\&+s\cdot(\epsilon_{\theta}(z_{\hat{P}}^t| I_{CLIP}, L)-\epsilon_{\theta}(z_{\hat{P}}^t|I_\emptyset,L_\emptyset)),
\end{split}
\end{align}
where \(\hat{\epsilon}_{\theta}\) represents the estimated noise, with \(s\) denoting the guidance scale and \(t\) indicating the timestep. \(I_\emptyset \) is predefined as fixed random noise, matching the dimensions of \(I_{CLIP}\), while \(L_\emptyset \) is assigned a constant value that signifies a non-specific type of apparel. This streamlined methodology substantially enhances the model's efficiency and elevates the quality of sampling.

\section{More details for SHHQe} \label{secA1}
\subsection{Data Statistics} 
To prevent potential data leakage, we partitioned the dataset into training and test sets based on the IDs of original images from SHHQ. This approach guarantees the same image in either training or testing, thus preventing any overlap. Consequently, the division of images for each apparel category naturally follows from the original IDs, resulting in the distribution of sample sizes for training and testing as detailed in Tab.~\(\ref{tab:statistics}\).

\begin{table}[h]
  \caption{Data Statistics of SHHQe.}
  \setlength{\tabcolsep}{1.8em}
  \label{tab:statistics}
  \centering
  \resizebox{\linewidth}{!}{
  \begin{tabular}{@{}lccc@{}}
    \toprule
    \textbf{Category} & \textbf{Train} & \textbf{Test} & \textbf{All} \\
    \midrule
    Tops  & 26,492  & 2,929 & 29,421 \\
    Pants  & 21,451  & 2,350 & 23,801 \\
    Dress  & 11,193  & 1,258 & 12,451\\
    Skirt  & 4,184  & 484 & 4,668 \\
    Shoes  &  34,797 & 3,863 & 38,660 \\
    Scarf  & 1,102  & 108 & 1,210 \\
    Bag  &  5,185  & 568 & 5753 \\
    Headwear  &  5,495 & 606 & 6,101 \\
    Socks  &  4,178 & 487 & 4,665 \\
    All  &  114,077 & 12,653 & 126,730 \\
  \bottomrule
  \end{tabular}}
  \vspace{-.4cm}
\end{table}

\subsection{Agnostic Images and Guidance Prompts}
\begin{figure*}[h]

    \centering
    \includegraphics[width=0.75\linewidth]{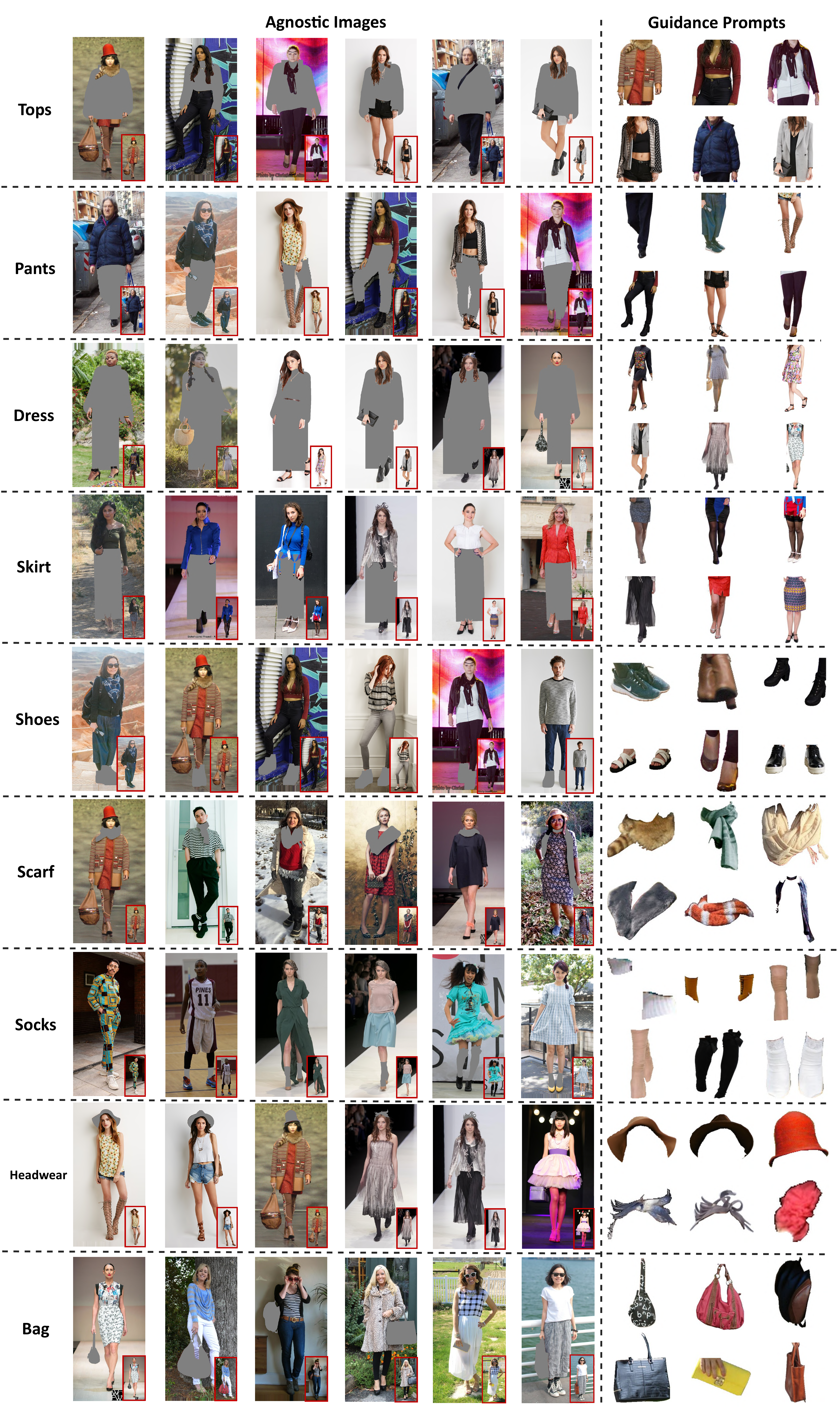}
    \caption{Agnostic Images and their corresponding Guidance Prompts, with the bottom left corner displaying the raw images.}
    \label{fig:agnostic_1}
\end{figure*}

Fig. \(\ref{fig:agnostic_1}\) shows agnostic images and guidance prompts extracted using the method described in Sec. \(3.1\), in which examples across nine types of apparel. Additionally, we utilize GPT-4 to generate unpaired textual descriptions for testing, as shown in Tab. \(\ref{tab:description}\). We use carefully designed prompts to generate various text descriptions that include style information such as color, texture, \textit{etc}.

\section{Discussion}
Our model supports both textual and style prompts for editing, but the latter differs from pixel-wise virtual try-on tasks~\cite{choi2021viton,xie2023gp,gou2023taming,han2018viton,lewis2021tryongan}. By leveraging aligned CLIP features, our approach prioritizes texture, material, and style transfer over precise pixel-level details. 

\begin{figure*}[h!]
    \centering
    \includegraphics[width=0.72\linewidth]{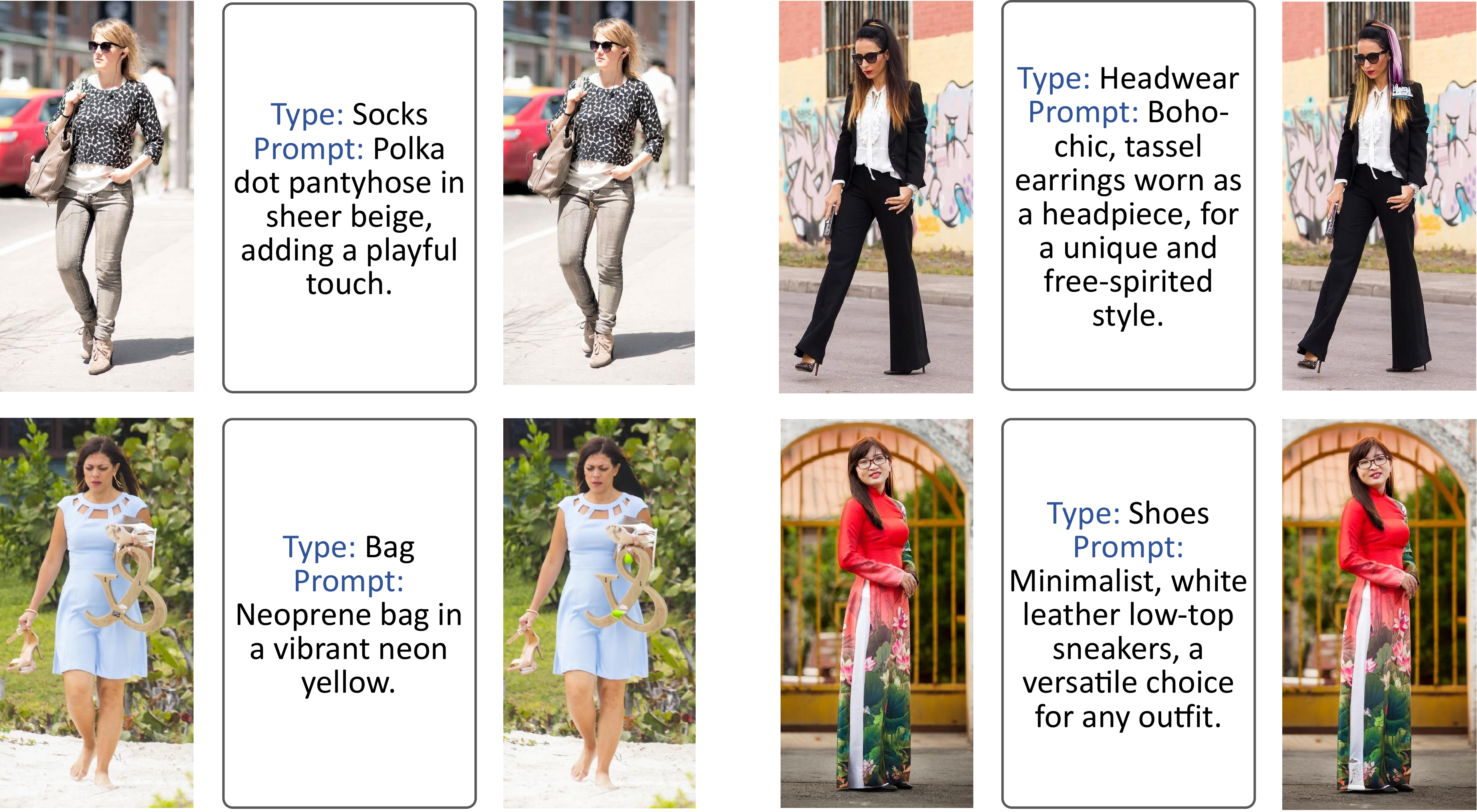}
    \caption{Examples of failure cases. Apparel with small areas and particularly complex scenes may lead to editing failures.}
    \label{fig:shhq_4}
\end{figure*}

We adopt a weak removal strategy for some categories of apparel, which may limit editing diversity. Moreover, editing apparel with rare styles or in small areas remains challenging, causing some failure cases as detailed in Fig.~\(\ref{fig:shhq_4}\). Future work will aim to improve the data extension method and the editing accuracy for small and peculiar apparel.

\begin{figure*}[h!]
    \centering
    \includegraphics[width=0.78\linewidth]{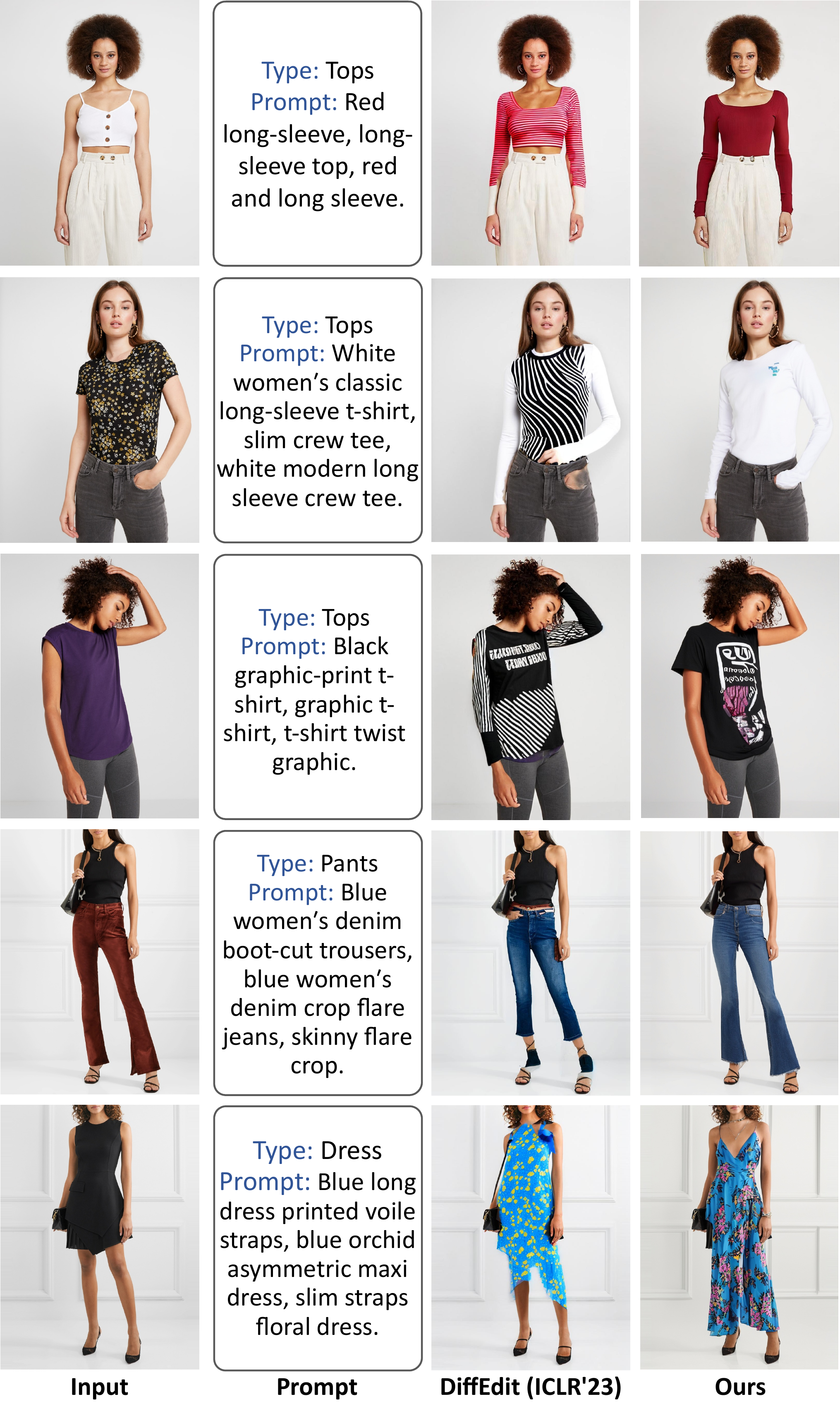}
    \caption{Visual Comparison of our model and DiffEdit.}
    \label{fig:compare_diffedit}
\end{figure*}

\clearpage

\clearpage

\clearpage

\clearpage

\clearpage

\clearpage

\end{document}